\documentclass[]{gPAA2e}

\newcommand{\sgn}{\operatorname{sgn}}
\newcommand{\sgnbar}{\overline{\operatorname{sgn}}}

\usepackage{algpseudocode}

\begin{document}
\doi{10.1080/17445760.YYYY.CATSid}
 \issn{1744-5779}
\issnp{1744-5760}
\jvol{00} \jnum{00} \jyear{2016} 

\markboth{M. Alexander Nugent and Timothy W. Molter}{Knowm Inc.}

\title{Thermodynamic-RAM Technology Stack}

\author{
M. Alexander Nugent$^{\rm a}$ and Timothy W. Molter$^{\rm a}$$^{\ast}$\thanks{$^\ast$Corresponding author. Email: tim@knowm.org
\vspace{6pt}}\\\vspace{6pt}  $^{\rm a}${\em{Knowm Inc., Santa Fe, NM, USA}}; 
}

\maketitle

\begin{abstract}
We introduce a technology stack or specification describing the multiple levels of abstraction and specialization needed to implement a neuromorphic processor (NPU) based on the previously-described concept of AHaH Computing and integrate it into today's digital computing systems. The general purpose NPU implementation described here is called Thermodynamic-RAM (kT-RAM) and is just one of many possible architectures, each with varying advantages and trade offs. Bringing us closer to brain-like neural computation, kT-RAM will provide a general-purpose adaptive hardware resource to existing computing platforms enabling fast and low-power machine learning capabilities that are currently hampered by the separation of memory and processing, a.k.a the von Neumann bottleneck. Because understanding such a processor based on non-traditional principles can be difficult, by presenting the various levels of the stack from the bottom up, layer by layer, explaining kT-RAM becomes a much easier task. The levels of the Thermodynamic-RAM technology stack include the memristor, synapse, AHaH node, kT-RAM, instruction set, sparse spike encoding, kT-RAM emulator, and SENSE server.

\bigskip


\bigskip

\end{abstract}

\section{Introduction}

Machine learning applications span a very diverse landscape. Some areas include motor control, combinatorial search and optimization, clustering, prediction, anomaly detection, classification, regression, natural language processing, planning and inference. A common thread is that a system learns the patterns and structure of the data in its environment, builds a model, and uses that model to make predictions of subsequent events and take action. The models which emerge contain hundreds to trillions of continuously adaptive parameters. Human brains contain on the order of $10^{15}$ adaptive synapses. How the adaptive weights are exactly implemented in an algorithm varies, and established methods include support vector machines, decision trees, artificial neural networks and deep learning, to name a few \cite{marsland2015machine}. Intuition tells us learning and modeling the environment is a valid approach in general, as the biological brain also appears to operate in this manner. The unfortunate limitation with the algorithmic approach, however, is that it runs on traditional digital hardware. In such a computer, calculations and memory updates must necessarily be performed in different physical locations, often separated by a significant distance. The power required to adapt parameters grows impractically large as the number of parameters increases owing to the tremendous energy consumed shuttling digital bits back and forth. In a biological brain (and all of nature), the processor and memory are the same physical substrate and many computations and memory adaptations are performed in parallel. Recent progress has been made with multi-core processors and specialized parallel processing hardware like GP-GPUs \cite{xiong2016achieving} and FPGAs \cite{lacey2016deep}, but for machine learning applications that intend to achieve the ultra-low power dissipation of biological nervous systems, it is a dead end approach \cite{potok2016neuromorphic}. The low-power solution to machine learning occurs when the memory-processor distance goes to zero, and this can only be achieved through intrinsically adaptive hardware, such as memristors.

Given the success of recent advancements in machine learning algorithms combined with the hardware power dilemma, an immense pressure exists for the development neuromorphic computer hardware. The Human Brain Project and the BRAIN Initiative with funding of over EUR 1.190 billion and USD 3 billion respectively partly aim to reverse engineer the brain in order to build brain-like hardware \cite{hampton2014european,insel2013nih}. DARPA's recent SyNAPSE program funded two large American tech companies IBM and HP as well as research giant HRL labs and aimed to develop a new type of cognitive computer similar to the form and function of a mammalian brain. The recent Nanotechnology-Inspired Grand Challenge for Future Computing in the United States \cite{chennanotechnology} was formed to "Create a new type of computer that can proactively interpret and learn from data, solve unfamiliar problems using what it has learned, and operate with the energy efficiency of the human brain." CogniMem is commercializing a k-nearest neighbor application specific integrated circuit (ASIC) for pattern classification, a common machine learning task found in diverse applications \cite{mccormick2014applying}. Stanford's Neurogrid, a computer board using mixed digital and analog computation to simulate a network, is yet another approach at neuromorphic hardware \cite{benjamin2014neurogrid}. Manchester University's SpiNNaker is another hardware platform utilizing parallel cores to simulate biologically realistic spiking neural networks\cite{navaridas2013spinnaker}. IBM's neurosynaptic core and TrueNorth cognitive computing system resulted from the SyNAPSE program \cite{esser2013cognitive}. All these platforms have yet to prove utility along the path towards mass adoption, and none have yet solved the foundational problem of memory-process separation.

More rigorous theoretical frameworks are also being developed for the neuromorphic computing field. Recently, Traversa and Ventra have introduced the idea of `universal memcomputing machines', a general-purpose computing machine that has the same computational power as a non-deterministic Universal Turing Machine showing intrinsic parallelization and functional polymorphism \cite{traversa2014universal}. Their system and other similar proposals employ a relatively new electronic component, the memristor, whose instantaneous state is a function of its past states. In other words, it has memory, and like a biological synapse, it can be used as a subcomponent for computation while at the same time storing a unit of data. A previous study by Thomas et al. demonstrated that the memristor can better be used to implement neuromorphic hardware than traditional CMOS electronics \cite{thomas2013memristor}.

Our attempt to develop neuromorphic hardware takes a unique approach inspired by life, and more generally, natural self-organization. We call the theoretical result of our efforts `AHaH Computing' and have previously provided a thorough and rigorous quantitative description \cite{nugent2014ahah}. Rather than trying to reverse engineer the brain or transfer existing machine learning algorithms to new hardware and blindly hope to end up with an elegant power efficient chip, AHaH computing was designed from the beginning with a few key constraints: (1) must result in a hardware solution where memory and computation are combined, (2) must enable most or all machine learning applications, (3) must be simple enough to build chips with existing manufacturing technology and emulated with existing computational platforms for verification of methods (4) must be understandable and adoptable by application developers across all manufacturing sectors. This initial motivation led us to utilize physics create a technological framework for a neuromorphic processor satisfying the above constraints. 

In trying to understand how nature computes, we stumbled upon a fundamental structure found not only in the brain but also almost everywhere one looks - a self-organizing energy-dissipating fractal. We find it in rivers, trees, lighting and fungus, but we also find it deep within us. The air that we breathe is coupled to our blood through thousands of bifurcating flow channels that form our lungs. Our brain is coupled to our blood through thousands of bifurcating flow channels that form our arteries and veins. The neurons in our brains are built of thousands of bifurcating flow channels that form our axons and dendrites. At all scales of organization we see the same fractal built from the same simple building block: a simple structure formed of competing energy dissipation pathways. We call this building block `nature's Transistor', as it appears to represent a foundational adaptive building block from which higher-order self-organized structures are built, much like the transistor is a building block for modern computing.

When multiple conduction pathways compete to dissipate energy through an adaptive container, the container will adapt in a particular way that leads to the maximization of energy dissipation. We call this mechanism the Anti-Hebbian and Hebbian (AHaH) plasticity rule. It is computationally universal, but perhaps more importantly and interestingly, it also leads to general-purpose solutions in machine learning. Because the AHaH rule describes a physical process, we can create efficient and dense analog AHaH synaptic circuits with memristive components. One version of these mixed signal (digital and analog) circuits forms a generic adaptive computing resource we call Thermodynamic Random Access Memory or Thermodynamic-RAM, described herein. Thermodynamics is the branch of physics that describes the temporal evolution of matter as it flows from ordered to disordered states, and nature's Transistor is an energy-dissipation flow structure, hence `thermodynamic'.

In neural systems, the algorithm is specified by two things: the network topology and the plasticity of the interconnections or synapses. Any general-purpose neural processor must contend with the problem that hard-wired neural topology will restrict the available neural algorithms that can be run on the processor. It is also crucial that the NPU interface merge easily with modern methods of computing. A `Random Access Synapse' structure satisfies these constraints.

Thermodynamic-RAM is the first attempt at realizing a working neuromorphic processor implementing the theory of AHaH computing. While several alternative designs, such as dual crossbars, are feasible and may offer specific advantages over others, this first design aims to be a general computing substrate geared towards reconfigurable network topologies and the entire spectrum of the machine learning application space. In the following sections, we break down the entire design specification into various levels from ideal memristors to integrating the finished product into existing technology. Defining the individual levels of this `technology stack' helps to introduce the technology step by step and group the necessary pieces into tasks with focused objectives. This allows for separate groups to specialize at one or more levels of the stack where their strengths and interests exist. Improvements at various levels can propagate throughout the whole technology ecosystem, from materials to markets, without any single participant having to bridge the whole stack. In a way, the technology stack is an industry specification.

\section{The Thermodynamic-RAM Technology Stack}

\subsection{The Memristor -- Metastable Switch Collection}

\begin{figure}
\begin{center}
\begin{minipage}{100mm}
\includegraphics[width=1.0\linewidth]{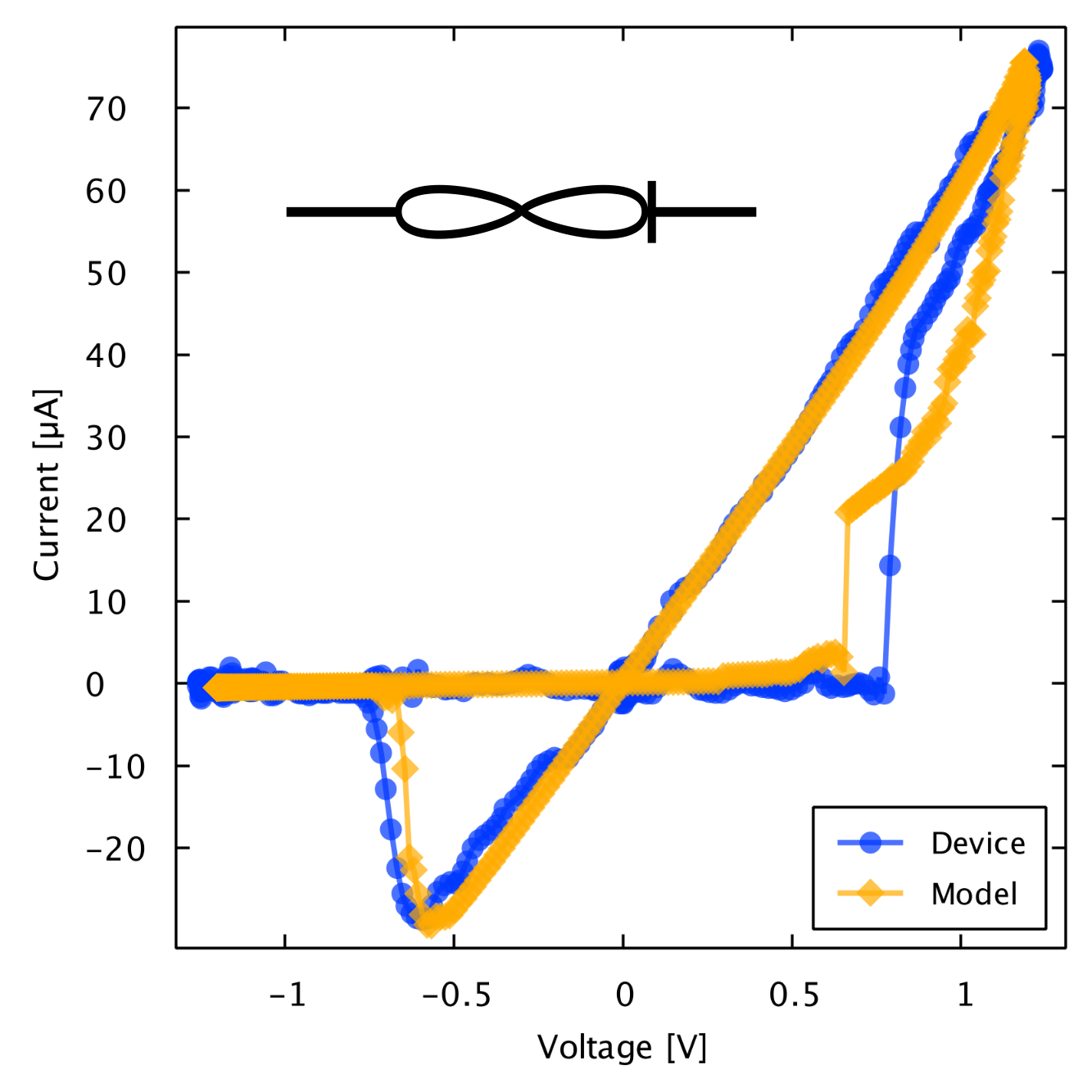}%
\caption{Our generalized memristor model captures both the memory and exponential diode characteristics via metastable switches (MSS) and parallel Schottky diodes and provides an excellent model for a wide range of memristive devices. Here we show a hysteresis plot for a Ag-chalcogenide device from Boise State University along with a fitted model.}%
\label{fig_memristor}
\end{minipage}
\end{center}
\end{figure}

Many memristive materials have recently been reported \cite{campbell2016self,oblea2010silver,yang2012complementary,valov2013cation,hasegawa2011memristive,jackson2013nanoscale}, and the trend continues. New designs and materials are being used to create a diverse range of devices. Memristor models are also being developed and incrementally improved upon \cite{choi2012resistance,menzel2012simulation,chang2011synaptic,sheridan2011device,biolek2009spice}. Our generalized metastable switch (MSS) memristor model is to date a candidate for the most accurate model shown to capture the behavior of memristors at a level of abstraction sufficient to enable efficient circuit simulations while simultaneously describing as wide a range of devices as possible \cite{nugent2014ahah}. A MSS is an idealized two-state element that switches probabilistically between its two states as a function mainly of the applied time-voltage integral. The MSS model describes a memristor as a \textit{collection} of MSSs evolving in time, which captures important device behavior such as hysteresis in response to an oscillating excitation and incremental conductance change in response to voltage pulses. The MSS model can be made more complex to account for failure modes, for example by making the MSS state potentials temporally variable. Multiple MSS models with different state potentials can be combined in parallel or series to model increasingly more complex state systems.

In our semi-empirical model, the total current through the device comes from both a memory-dependent (MSS) current component, $I_{\rm m}$, and a Schottky diode current, $I_{\rm s}$ in parallel: 

\begin{equation}
\label{mss_current}
I=\phi I_{\rm m}(V,t)+(1-\phi)I_{\rm s}(V),
\end{equation}

where $\phi\in{[0,1]}$. A value of $\phi=1$ represents a device that contains no Schottky diode effects. Tuning $\phi$ more towards zero gradually introduces more diode current while scaling back the memory component. The Schottky diode effect accounts for the exponential behavior found in many memristor devices made of sandwiched layers of metal and semiconductor material and allows for the accurate modeling of that effect, which the MSS component cannot capture alone. 

Thermodynamic-RAM is not constrained to just one particular memristive device; any memristive device can be used as long as it meets the following criteria: (1) it is incremental and (2) its state change is voltage dependent. The ideal device for neuromemristive processors would have low thresholds of adaptation (\textless 0.2~V, reduce power loses  during learning $V^2\cdot R$), on-state resistance of $\sim$10~k$\Omega$ or greater (reduce static current loses $P=I^2/R$), high dynamic range (increase synaptic weight resolution), durability (increase life of chip), the capability of incremental operation with very short pulse widths (increase learning precision and reduce energy $E=P * \Delta{t}$) and long retention times of a week or more (reduce loss of trained state). However, even devices that deviate considerably from these parameters will be useful in more specific applications. For example, short retention times on the order of seconds are perfectly compatible with combinatorial optimizers.

We have previously shown that our generalized MSS model for memristors accurately models four potential memristor candidates \cite{nugent2014ahah} for Thermodynamic-RAM, and we have incorporated the model into our circuit simulation and machine learning benchmarking software. A recent Ag-chalgogenide memristor device fabricated by Kris Campbell at Boise State University and model hysteresis plot is shown in Figure \ref{fig_memristor}. The model provides common ground from which the diversity of devices can be compared and incorporated into the technology stack. By modeling a device with the MSS model, a material scientist can evaluate its utility across real-world benchmarks via software emulators and gain valuable insight into which memristive properties are, and are not, useful in the application space.

\subsection{The Synapse -- Competing Memristors}

\begin{figure}
\begin{center}
\begin{minipage}{100mm}
\includegraphics[width=1.0\linewidth]{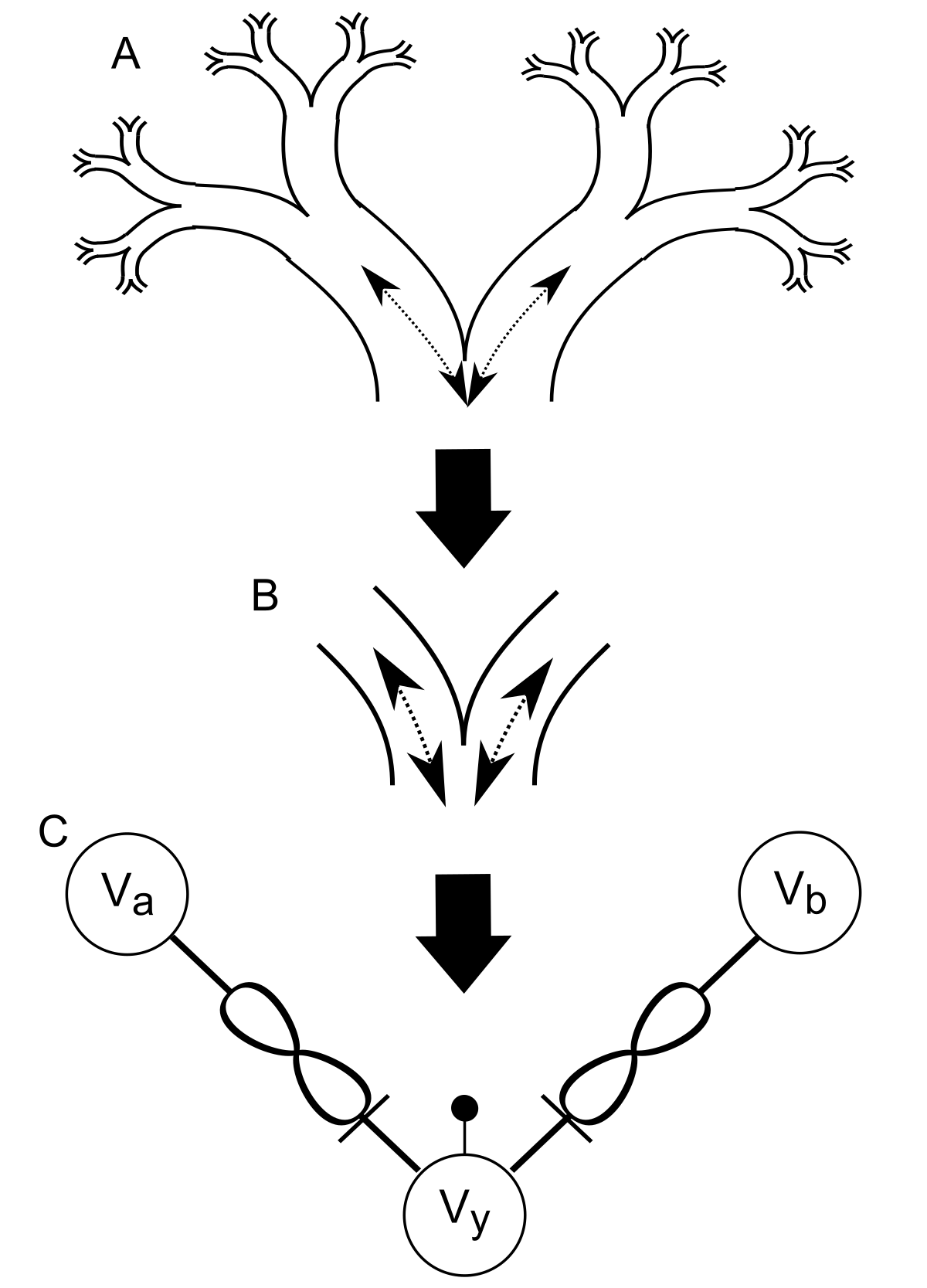}%
\caption{A) A self-organizing energy-dissipating fractal can be found throughout nature, is composed of a simple repeating structure formed of competing energy dissipation pathways. B) The simple bifurcating dissipative pathway is what we call nature's transistor or synapse. C) A differential pair of memristors provide a means for implementing a synapse in our electronics.}%
\label{fig_kt_synapse}
\end{minipage}
\end{center}
\end{figure}

As a variable conductance device, a memristor is an adaptive energy-dissipating pathway. As current flows through it, its internal state changes and heat is exchanged to the surrounding environment. When two adaptive energy-dissipating pathways compete for conduction resources, a nature's transistor will emerge. Two competing memristors thus form a synapse as shown in Figure \ref{fig_kt_synapse}. We see this building block for self-organized structures throughout nature, for example in arteries, veins, lungs, neurons, leaves, branches, roots, lightning, rivers and mycelium networks. We observe that in all cases there is a particle that flows through competitive energy-dissipating assemblies. The particle is either directly a carrier of free energy dissipation or else it appears to gate access, like a key to a lock, to free energy dissipation of the units in the collective. Some examples of these particles include water in plants, ATP in cells, blood in bodies, neurotrophins in brains, and money in economies. In the cases of whirlpools, hurricanes, tornadoes and convection currents we note that although the final structure does not appear to be built of competitive structures, it is the result of a competitive process with one winner; namely, the spin or rotation.

The circuits capable of achieving AHaH plasticity can be broadly categorized by the electrode configuration that forms the synapse as well as how the input activation (current) is converted to a feedback voltage that drives unsupervised anti-Hebbian learning \cite{nugent2008universal,nugent2009method}. Synaptic currents can be converted to a feedback voltage statically (resistors or memristors), dynamically (capacitors), or actively (operational amplifiers). Each configuration requires unique circuitry to drive the electrodes so as to achieve AHaH plasticity, and multiple driving methods exist. Non-Polar, Polar, and Bipolar memristors can be used. These are defined as following:

\begin{itemize}
\item Non-Polar: Application of both positive and negative voltage bias induces only increase or only decrease in conductance. Thermodynamic decay is used to change the conductance in the other direction. Examples include the dielectrophoretic aggregation of conductive nanoparticles in colloidal suspension \cite{wissnerGross2009DEP} . 
\item Polar: Application of voltage bias enables incremental conductance in one direction, but all-or-nothing change in the opposite direction. An example of this includes phase-change memory \cite{burr2017IBM} .
\item Bi-Polar: Application of positive and negative voltage bias enables incremental conductance increase and decrease. An example of this include Self Directed Channel (SDC) memristors \cite{campbell2016self} . 
\end{itemize}

All of these devices can be used as adaptive dissipation pathways and, via a specialized circuit, be made to compete for conduction resources. Hence, a large number of AHaH circuits exist. Herein, a `2-1' two-phase circuit configuration with polar memristors is introduced because of its compactness and because it is amenable to simple mathematical analysis \cite{nugent2014ahah}.

\subsection{The AHaH Node -- Collections of Synapses}

\begin{figure}
\begin{center}
\begin{minipage}{100mm}
\includegraphics[width=1.0\linewidth]{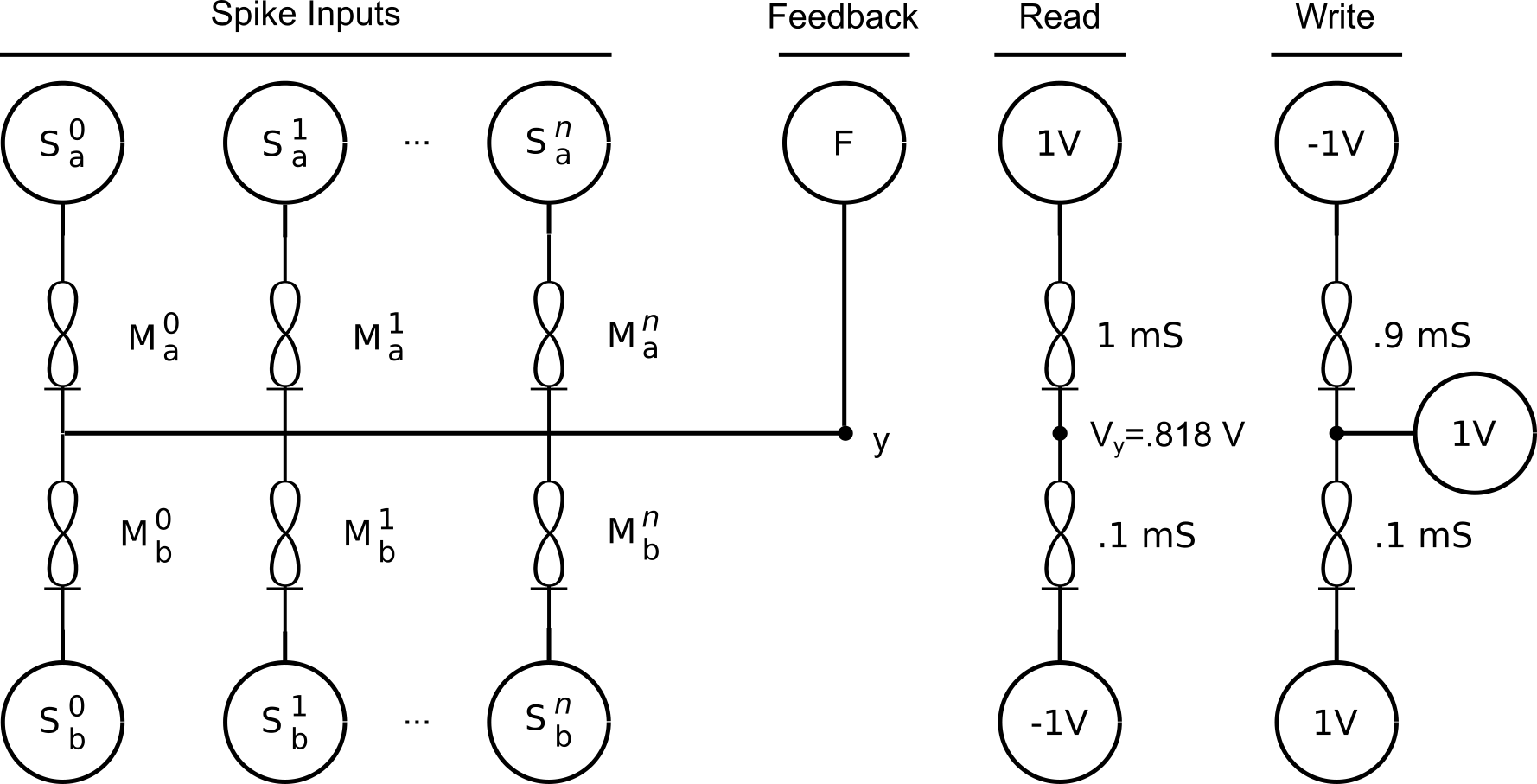}%
\caption{An AHaH node is made up of $n$ synapses sharing a common output electrode, $y$. The memristor pair synapse and the AHaH node are analogous to a biological synapse and neuron, respectively. In Thermodynamic-RAM, the number of input synapses can be configured via software and several AHaH nodes can be connected together to form any desired network topology by a technique called temporal partitioning.}%
\label{fig_ahah_node}
\end{minipage}
\end{center}
\end{figure}

An AHaH node is formed when a collective of synapses are coupled to a common readout line. Through spike encoding and temporal multiplexing, an AHaH node is capable of being partitioned into smaller functional AHaH nodes. An AHaH node provides a simple but computationally universal (and extremely useful) adaptation resource.

The functional objective of the AHaH node shown in Figure~\ref{fig_ahah_node} is to produce an analog output signal on electrode $y$, given an arbitrary spike input of length $n$ with $k$ active inputs and $n-k$ inactive (floating) inputs. The circuit consists of one or more memristor pairs (synapses) sharing a common electrode labeled $y$. Switches gating access to a driving voltage are labeled with an $S$. The individual switches for spike inputs of the AHaH node are labeled ${\rm S}^0$, ${\rm S}^1$ $\cdots$ ${\rm S}^n$. The driving voltage source for supervised and unsupervised learning is labeled $F$. The subscript values $a$ and $b$ indicate the positive and negative dissipative pathways, respectively.

Similar to a binary perceptron \cite{rosenblatt1958perceptron}, an AHaH node is able to linearly separate two classes of samples represented as n-dimensional input vectors. The output is the sum of the products of all inputs and weights plus a bias. Unlike a perceptron, an AHaH node is not an algorithm but an adaptive hardware construct, i.e. a physical adapting circuit where the output is an analog value representing the sum or integration of currents of physical synapse circuits cable of Anti-Hebbian and Hebbian plasticity. The synaptic weight is the difference in conductance between the differential memristor pair.

During the read phase, switches ${\rm S_a}$ and ${\rm S_b}$ connect voltage sources $+V$ and $-V$ respectively for all $k$ active inputs. Inactive $S$ inputs are left floating. The combined conductance of the active inputs produce an output voltage on electrode $y$. This analog signal contains useful confidence information and can be digitized via the ${\sgn()}$ function to either a logical 1 or a 0, if desired. The read example in Figure~\ref{fig_ahah_node} shows a simple case with one active synapse, $V=1 V$, $M_a=1 mS$ and $M_b=0.1 mS$. The resulting voltage divider circuit produces a voltage $V_y$, which in this case is 0.818 V. The act of reading also decays the synaptic weight slightly towards 0 V. This is the anti-Hebbian part of the AHaH rule.

During the write phase, voltage source $F$ is set to either $V_{\rm y}^{\rm write} = V \sgnbar{\left( {{V_{\rm y}}^{\rm read}} \right)}$ (unsupervised) or $V_{\rm y}^{\rm write} = V \sgnbar{(s)}$ (supervised), where $s$ is an externally applied teaching signal. The polarity of the driving voltage sources gates by the switches ${S}$ are inverted to $-V$ and $+V$. The polarity switch causes all active memristors to be driven to a less conductive state, counteracting the read phase. If this dynamic counteraction did not take place, the memristors would quickly saturate into their maximally conductive states, rendering the synapses useless. The write example in Figure~\ref{fig_ahah_node} extends the simple read example, where a 1V teaching signal is applied at node $y$. This applied voltage will case a -2 V drop across $M_a$, driving it into a less conductive state. The actual ending conductance value will depend on the time the voltage is applied. This is the Hebbian part of the AHaH rule.

A more intuitive explanation of the above feedback cycle is that ``the winning pathway is rewarded by not getting decayed.'' Each synapse can be thought of as two competing energy dissipating pathways (positive or negative evaluations) that are building structure (differential conductance). We may apply reinforcing Hebbian feedback by (1) allowing the winning pathway to dissipate more energy or (2) forcing the decay of the losing pathway. If we chose method (1) then we must at some future time ensure that we decay the conductance before device saturation is reached. If we chose method (2) then we achieve both decay and reinforcement at the same time. Method (2) is faster while method (1) is more energy efficient. The lowest energy solution is to use natural decay rather than forced decay, but this introduces complexities associated with matching the decay rate to the particular processing task.

\subsection{Sparse Spike Encoding -- Information Encoding}

\begin{figure}
\begin{center}
\begin{minipage}{100mm}
\includegraphics[width=1.0\linewidth]{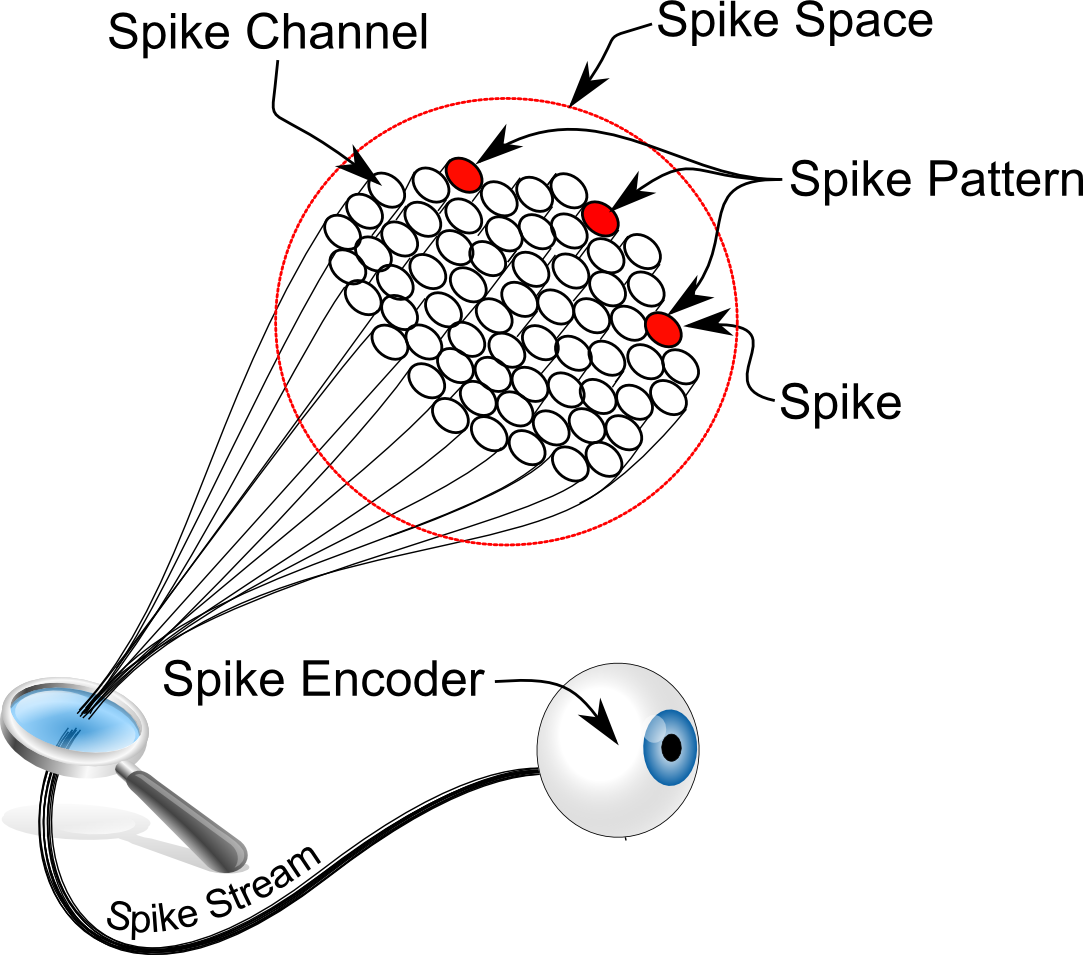}%
\caption{A spike-based system such as kT-RAM requires spike encoders (sensors), spike streams (wire bundles), spike channels (a wire), spike space (number of wires), spike patterns (active spike channels) and finally spikes (the state of being active). A spike encoding is, surprisingly, nothing more than a list of encoders that directly address synapses on a kT-RAM core.}%
\label{fig_spikes}
\end{minipage}
\end{center}
\end{figure}

A spike stream is the means in which real-world data is asynchronously fed into kT-RAM. Its biological counterpart would be the bundles of axons of the nervous system which carry sensed information from sensing organs to and around the cortex. A sparse spike stream interface is the only option with kT-RAM, and it is used for all machine learning applications from robotic control to clustering to classification. This trait enables an application developer to leverage their knowledge and experience using kT-RAM in one domain and transfer it over to another. Spikes can directly address core synapses. The synaptic core address can thus be given by the sum of the AHaH node's core partition index and the spike ID, which are both just integers in the spike space. Spikes enable kT-Core partitioning and multiplexing, which in turn enables arbitrary AHaH node sizes and hence very flexible network topologies. Sparse spike encoding is also very energy and bandwidth efficient and has shown to produce state-of-the-art results on numerous benchmarks. We choose spikes because they work, and we are attempting to engineer a useful computing substrate. The fact that the spike encoding appears to match biology is of course curious, but ultimately not important to our objectives.

A collection of $n$ synapses belong to a neuron (AHaH node), each with an associated weight:
$\{ w_0, w_1, \cdots w_n\}$. A subset of the synapses in an AHaH node can be activated by some input spike pattern, and the total neural activation is the voltage of the H-Tree, which can be read out on the common electrode, $y$ by the AHaH Controller. For many input patterns, $x$ is a sparse spiking representation, meaning that only a small subset of the spike channels are activated out of the spike space, and when they are, they are of value 1. So for a neuron with 16 inputs, one possible sparse-spike pattern would look like: $x=\{1000001000000000\}$. Since two of the 16 possible inputs are active (spiking), we say that it has a sparsity of $2/16$ or 12.5\%. Since most of the inputs are zero, we can write this spike pattern in a much more efficient way by just listing the index of the inputs that are spiking: $x=\{0,6\}$.

We call $x$ a `spike set' or `spike pattern' or sometimes just `spikes'. The `spike space' is the total number of `spike channels', in this case 16. In some problems such as inference or text classification the spike space can get all the way up to 250,000 or more. A good way to picture it is as a big bundle of wires, where the total number of wires is the spike space and the set of wires active at any given time is the spike pattern. We call this bundle of wires and the information contained in it the `spike stream'. The algorithms or hardware that convert data into a sparse-spiking representation are called `spike encoders'. Your eyes, ears and nose are examples of spike encoders. A visual representation of this can be seen in Figure~\ref{fig_spikes}.

\subsection{kT-RAM -- AHaH Nodes with RAM Interface}

\begin{figure}
\begin{center}
\begin{minipage}{100mm}
\includegraphics[width=1.0\linewidth]{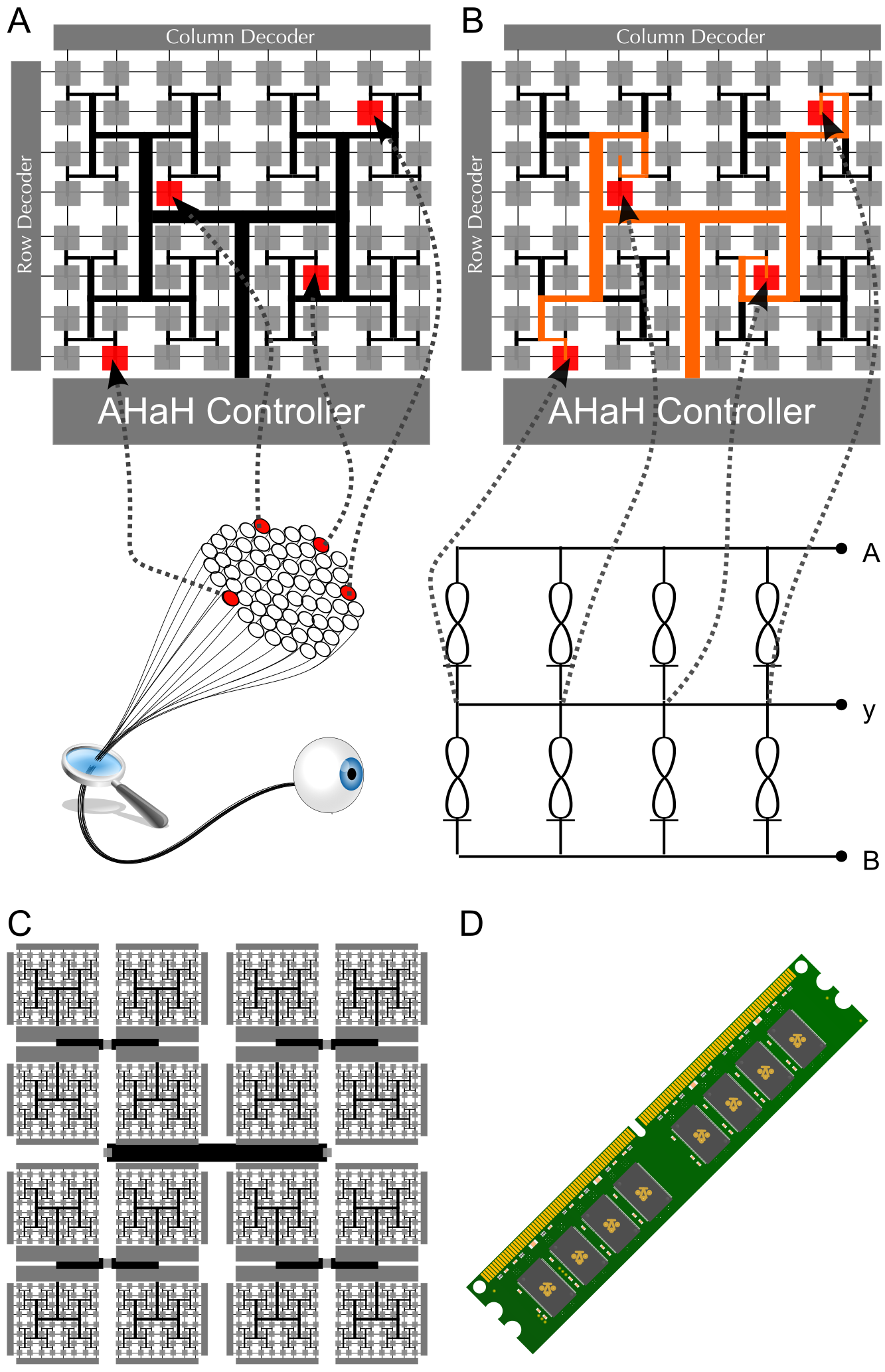}%
\caption{A) Spikes (integers) in a spike pattern (integer set) are used to address synaptic elements in a core, which become selectively coupled to drive circuitry (AHaH Controller). B) During the read and write phases, the activated synapses (memristor pairs) are coupled to the triple H-tree electrodes $A$, $B$ and $y$. C) By coupling several cores together, via either analog or digital methods, large collections or core (kT-RAM) can be created and specialized for tasks such as high-dimension inference (analog coupling) or compositional learning (digital coupling). D) kT-RAM can borrow from existing RAM architecture to integrate into existing digital computing platforms.}%
\label{fig_kt_ram}
\end{minipage}
\end{center}
\end{figure}

As previously stated, the particular design of kT-RAM presented in this paper prioritizes flexibility and general utility above anything else, much in the same way that a CPU is designed for general purpose use. This particular design builds upon commodity RAM using its form factor and the row and column address space mapping to specific bit cells. Modifying RAM to create a kT-RAM core requires the following steps: (1) removal of the RAM reading circuitry, (2) minor design modifications of the RAM cells, (3) the addition of memristive synapses to the RAM cells, (4) addition of H-Tree circuitry connecting the synapses, (5) and addition of driving and output sensing circuitry - the `AHaH controller'. Multiple kT-RAM cores can be manufactured and connected to each other on the same die (Figure~\ref{fig_kt_ram}~C). Leveraging existing techniques and experience of foundries capable of producing commodity RAM as well as using three to five generation-old processing facilities will make the prototyping and manufacturing of kT-RAM relatively inexpensive. Even the final packaging of kT-RAM modules (Figure~\ref{fig_kt_ram}~D) can leverage existing commodity hardware infrastructure.

Figure~\ref{fig_kt_ram}A and B show what kT-RAM would look like with its triple H-Tree sensing node connecting all the underlying synapses located at each cell in the RAM array. The 3 fractal binary trees shown (although from above it look like one single H-tree there are 3 stacked on top of each other) are the AHaH node's output electrode, $y$, as well as the driving voltage sources $A$ and $B$ connected via the switches as shown in Figure~\ref{fig_ahah_node}. The switches are activated via the incoming spikes (integers) and couple the individual synapses (memristor pairs) to the triple H-tree in preparation for the read and write cycles. All other synapses in the H-tree are left floating i.e. not connected to the triple H-tree. While at first glance it appears like this architecture leads to one giant AHaH node per chip or core, the core can be partitioned into smaller AHaH nodes of arbitrary size by temporally partitioning sub portions of the tree. In other words, so long as it is guaranteed that synapses assigned to a particular AHaH node partition are never co-activated with other partitions, these `virtual' AHaH nodes can co-exist on the same physical core. This allows us to effectively exploit the extreme speed of modern electronics. Any desired network topology linking AHaH nodes together can be achieved easily through a kT-RAM/CPU/RAM paring. Software enforces the constraints, while the hardware remains flexible. 

Through temporal partitioning combined with spike encoding, AHaH nodes can be allocated with as few as one or as many synapses as the application requires and can be connected to create any network topology. This flexibility is possible because of a RAM interface with addressable rows and columns. Crossbar architectures, in addition to sneak-path issues, introduce a restrictive topology. While this is good for specialized applications, one cannot build a general-purpose machine learning substrate from an intrinsically restricted topology. Cores can be electrically coupled to form a larger combined core. The number of cores, and the way in which they are addressed and accessed will vary across implementations so as to be optimized for end use applications. AHaH node sizes can therefore vary from one synapse to the size of the kT-RAM chip, while digital coupling could extend the maximal size to `the cloud', limited only by the kT-Core’s intrinsic adaptation rates and chip-to-chip communication.

\subsection{kT-RAM Instruction Set}

Thermodynamic RAM performs an analog sum of currents and adapts physically, eliminating the need to compute and write memory updates. One can theoretically exploit the kT-RAM instruction set (Table \ref{instruction_set}) however they wish. However, to prevent weight saturation, one must pair `forward' instructions with `reverse' instructions, although not necessarily one right after the other. For example, a forward-read operation $FF$ should be followed by a reverse operation ($RF$, $RH$, $RL$, $RZ$, $RA$ or $RU$) and vise versa. The only way to extract state information is to leave the feedback voltage floating, and thus there are two possible read instructions: $FF$ and $RF$. There is no such thing as a `non-destructive read' operation in kT-RAM, as this property is governed by the underlying physics of the memristive elements, which can differ. Every memory access results in weight adaptation, although it should be noted that operating under the adaptation threshold of some devices may result in negligible changes. By understanding how the Anti-Hebbian and Hebbian plasticity works (AHaH computing), we can exploit weight adaptations to create, among other things, `self-healing hardware'. The act of accessing the information actually repairs and heals it. 

\begin{table}
\tbl{kT-RAM Instruction Set}
{\begin{tabular}{@{}lcccccc}\toprule
Instruction & Synapse Driving Voltage & Feedback Voltage (F)\\
\colrule
FF & Forward-Float & None/Floating \\
FH & Forward-High & $-V$ \\
FL & Forward-Low & $+V$ \\
FU & Forward-Unsupervised & $-V$ if $y\geq0$ else $+V$ \\
FA & Forward-Anti-Unsupervised & $+V$ if $y\geq0$ else $-V$ \\
FZ & Forward-Zero & $0$ \\
RF & Reverse-Float & None/Floating \\
RH & Reverse-High & $-V$ \\
RL & Reverse-Low & $+V$ \\
RU & Reverse-Unsupervised & $-V$ if $y\geq0$ else $+V$ \\
RA & Reverse-Anti-Unsupervised & $+V$ if $y\geq0$ else $-V$ \\
RZ & Reverse-Zero & $0$ \\
\botrule
\end{tabular}}
\label{instruction_set}
\end{table}

Figure \ref{fig_classifier} contains pseudo code demonstrating how to construct a multi-label online classifier by loading spikes and executing instructions in the kT-RAM instruction set. The network topology of the classifier is simply $N$ AHaH nodes with $M$ synapses, where $N$ is the number of labels being classified and $M$ is the number of unique spikes in the entire spike stream space. The active spikes $S$, a subset of $M$, is loaded onto each AHaH node, and the execute method returns the voltage on the AHaH node's output electrode, $y$. Although all the AHaH nodes may exist on the same physical chip and share the same output electrode, temporal partitioning, as described above, allows for a virtual separation of AHaH nodes. Note that the \textit{execute} method actually receive two instructions at the same time in the implementation of the execution set in order to enforce executing `reverse' instructions with `forward' instructions as discussed above. For the rare case where only one instruction should be executed, for example a `FF' read instruction, a special "do nothing" instruction is defined: `XX'"

\begin{figure}
\begin{algorithmic}[1]
\Procedure{Classify}{active spikes set $S$, truth labels set $L$}
\For{Each AHaH Node $N$}
\State KTRAM.loadSpikes($N$, $S$)
\State $y\gets$ KTRAM.execute($N$, $FF$) \Comment{forward read}
\If{supervised}

\If{$N \in L$}
\State KTRAM.execute($N$, $RH$)
\ElsIf{$y \geq 0$}\Comment{false-positive}
\State KTRAM.execute($N$, $RL$)
\Else \Comment{true-negative}
\State KTRAM.execute($N$, $RF$) 
\EndIf

\Else \Comment{unsupervised}
\State KTRAM.execute($N$, $RF$) 
\EndIf
\EndFor
\EndProcedure
\end{algorithmic}
\caption{A multi-label online linear classifier with confidence estimation can be easily constructed via calls to the kT-RAM instruction set.}
\label{fig_classifier}
\end{figure}

The Mixed National Institute of Standards and Technology (MNIST) database \cite{lecun1998mnist} is a classic dataset in the machine learning community. It consists of 60,000 train and 10,000 test samples of handwritten digits, each containing a digit 0 to 9 (10 classes). The 28 x 28 pixel grayscale images have been preprocessed to size-normalize and center the digits. This very basic test benchmark is commonly used in the machine learning community to test out new ideas and algorithms and numerous results are published. When using our kT-RAM emulator to run the pseudo-code shown in Figure \ref{fig_classifier} on the MNIST benchmark dataset, it produces an accuracy of 92.1\% congruent with state-of-the-art linear classifier algorithms. In order to improve on this result, additional machine learning techniques need to be employed beyond linear classification.

\subsection{kT-RAM Emulator -- Cross-platform Universality}

Thermodynamic-RAM is designed to plug into existing computing architectures easily. The envisioned hardware format is congruent with standard RAM chips and RAM modules and would plug into a motherboard in a variety of different ways. In general there are two main categories of integration. First, kT-RAM can be tightly coupled with the CPU, on the CPU die itself or connected via the north bridge. In this case, the instruction set of the CPU would have to be modified to accommodate the new capabilities of kT-RAM. Secondly, kT-RAM is loosely coupled as a peripheral device either via the PCI bus, the LPC bus, or via cables or ports to the south bridge. In these cases, no modification to the CPU's instruction set would be necessary, as the interfacing would be implemented over the generic plug-in points over the south bus. As in the case with other peripheral devices, a device driver would need to be developed. Additional integration configurations are also possible.

Given the envisioned hardware integration, kT-RAM simply becomes an additional resource that software developers have access to via an API. In the meantime, kT-RAM is implemented as an emulator running on von Neumann architecture (for more machine learning benchmarks see \cite{nugent2014ahah}), but the API will remain the same. Later, when the new NPU is available, it will replace the emulator, and existing programs will not need to be rewritten to benefit from the accelerated capabilities offered by the hardware. In any case, kT-RAM operates asynchronously. As new spike streams arrive, the driver in control of kT-RAM is responsible for activating the correct synapses and providing the AHaH controller with an instruction pair for each AHaH node. The returned activation value can then be passed back to the program and used as needed. 

Emulators allow developers to commence application development while remaining competitive with competing machine learning approaches. In other words, we can build a market for kT-RAM across all existing computing platforms while we simultaneously build the next generation of kT-RAM hardware. kT-RAM software emulators for both memristive circuit validation and near-term application development on digital computers have already been developed and deployed commercially on real-world client problems. Our current digital kT-Core emulators seem to be extremely efficient running on commodity hardware compared to existing methods in performance, energy and memory efficiency, and a thorough comparison is planned for the near future. Thermodynamic-RAM is not a `ten year technology' nor is it `bleeding edge'. Rather, it is already solving real-world machine learning problems on existing digital platforms.

\subsection{SENSE Server -- Plug-and-Play Machine Learning Apps}

While a machine learning application developer using the kT-RAM emulator would have full control of the design of the application and can use kT-RAM to its full potential, she would be required to understand the instruction set and underlying mechanics of kT-RAM and AHaH Computing. This level of development is analogous to writing assembly code or using a very low-level programming library. 

To assist in the rapid development of applications based on kT-RAM, we have developed a top-level server-based framework. We call it `Scalable and Extensible Neural Sensing Engine' or `SENSE server' for short. The SENSE server contains higher level pre-built machine learning modules, standard spike encoders, buffers, spike stream joiners and other miscellaneous building blocks, which can be configured by the developer for a unique machine learning application. This level of development is analogous to an SQL server like MySQL, where you provide a configuration file to specify its behavior. Like the MySQL server, the SENSE server runs as a daemon service, waiting for asynchronous interactions from the outside world. In the case of the SENSE server, it is waiting for incoming spikes flowing in over the configured spike streams. To install and run the SENSE server on Linux, you would run a command in a terminal such as `sudo apt-get install knowm-sense' followed by `start knowm-sense myconfig.yml', where `myconfig.yml' would be the custom configuration file defining the `netlist' and parameter settings of the particular machine learning application. The SENSE server can be run on commodity computer hardware, robotic platforms or mobile devices with a Linux or *nix-based operating system. The SENSE server also has built-in support for seamless clustering for parallelization of high throughput machine learning applications such as vision and audio processing.

\section{Conclusion}

In this paper, we have introduced Thermodynamic-RAM and a technology stack, a specification or blueprint, for a future industry enabled by AHaH Computing. kT-RAM is a particular design that prioritizes flexibility and general utility above anything else, much in the same way that a CPU is designed for general purpose use. The flexibility offered by this design allows for a single architecture that can be used for the entire range of machine learning applications given their unique network topologies. Much like the cortex integrates signals from different sensing organs via a common `protocol', the sparse spike encoding interface of kT-RAM allows for a well defined way to integrate environmental data asynchronously. Conveniently, the sparse spike encoding interface is a perfect bridge between digital systems and neuromorphic hardware. Just as modern computing is based on the concept of the ‘bit’ and quantum computing is based on the concept of the ‘qubit’, AHaH computing is built from the ‘ahbit’. AHaH attractor states are a reflection of the underlying statistics (history) of the applied data stream. It is both the collection of physical synapses and also the structure of the information that is being processed that together result in an AHaH attractor state. Hence, an ‘ahbit’ is what results when we couple information to energy dissipation. Our kT-RAM design borrows heavily from commodity RAM using its form factor to build upon and leverage today's chip manufacturing resources. The RAM module packaging and concise instruction set will allow for easy integration into existing computing platforms such as commodity personal computers, smart phones and super computers. Our kT-RAM emulator allows us to develop applications, demonstrate utility, and justify a large investment into future chip development. When chips are available, existing applications using the emulator API will not have to be rewritten in order to take advantage of new hardware acceleration capabilities. The topmost level of the kT-RAM technology stack is the SENSE server, a framework for configuring a custom machine learning application, based on a `netlist' of pre-built machine learning modules, standard spike encoders, buffers, spike stream joiners and other miscellaneous building blocks. 

\section{Future Work}

At the core of the adaptive power problem is the energy wasted during memory–-processor communication. The ultimate solution to the problem entails finding ways to let memory configure itself, and AHaH computing is a conceptual framework for understanding how this can be accomplished. Thermodynamic-RAM is an adaptive physical hardware resource for providing AHaH plasticity and hence a substrate from which AHaH computing is possible. In previous work, we have shown demonstrations of universal logic, clustering, classification, prediction, robotic actuation and combinatorial optimization benchmarks using AHaH computing, and we have successfully mapped all these functions to the kT-RAM instruction set and emulator. Efficient emulation has already been demonstrated on commodity von Neumann hardware, and a path ahead towards neuromorphic chips has been defined. Along the way, the emulator will be ported to co-processors like GP-GPUs and FPGAs to further improve speed and power efficiency with available hardware. Progress is being made independently at various levels, but a coordinated and focused effort by multiple participants is needed to bridge the full technology stack. 

\section*{Acknowledgment}

The authors would like to thank the Air Force Research Labs in Rome, NY for their support under the SBIR/STTR programs AF10-BT31, AF121-049. The authors would like to thank Kristy A. Campbell from Boise State University for graciously providing us with memristor device data.
\vspace{12pt}
\bibliographystyle{gPAA}
\bibliography{2016_JPEDS}
\markboth{Taylor \& Francis and I.T. Consultant}{The International Journal of Parallel, Emergent and Distributed Systems}
 \vspace{36pt}

\label{lastpage}

\end{document}